\documentclass[letterpaper, 10 pt, conference]{ieeeconf}  % Comment this line out if you need a4paper

\IEEEoverridecommandlockouts                              % This command is only needed if 
\usepackage{graphics} % for pdf, bitmapped graphics files
\usepackage[pdftex]{graphicx}
\DeclareGraphicsExtensions{.pdf,.jpg,.png,.gif}
	\usepackage{amssymb}
	\usepackage{amsmath}
	\usepackage{array}
	\usepackage{trfsigns}%->FT Zeichen
	\usepackage{colonequals}
	\DeclareMathOperator*{\argminA}{arg\,min} % Jan Hlavacek
	\usepackage{bm}
    \newcommand{\uvec}[1]{\boldsymbol{\hat{{#1}}}}
    \newcommand\norm[1]{\left\lVert#1\right\rVert}

\usepackage[english]{babel}
\usepackage{cite}
\usepackage{breakcites}
\usepackage{float}

\makeatletter
\let\MYcaption\@makecaption
\makeatother

\usepackage[font=footnotesize]{subcaption}

\makeatletter
\let\@makecaption\MYcaption
\makeatother

\usepackage{placeins}
\newcommand{\squeezeup}{\vspace{-2.5mm}}
	\usepackage[pdftex,breaklinks=true]{hyperref}
	
	\newcommand*{\shortautoref}[1]{%
  \begingroup
    \def\sectionautorefname{Sec.}%
    \def\subsectionautorefname{Sec.}%
    \def\figureautorefname{Fig.}%
    \def\equationautorefname{Eqn.}%
    \autoref{#1}%
  \endgroup
}

\usepackage[nameinlink,capitalize]{cleveref}

\usepackage{listings}
\usepackage{xcolor}
\usepackage{stackengine}
\usepackage{threeparttable}

\newif\ifshow
\showfalse

\usepackage{xcolor}
\usepackage{color}
\newcommand\hl[1]{
  \ifmmode
    \begingroup
      \color{red}%
      #1%
    \endgroup
  \else
    \textcolor{red}{#1}%
  \fi
}

\usepackage{environ}
\NewEnviron{myhl}{
  \ifshow
  \leavevmode\color{red}\ignorespaces{\BODY{}
  }%
  \else
  \BODY{}
  \fi
}

\ifshow
\else
\renewcommand\hl[1]{#1}
\fi

\title{\LARGE \bf
Distributed Timed Elastic Band (DTEB) Planner: \\
Trajectory Sharing and Collision Prediction for Multi-Robot Systems
}

\author{Yiu Ming Chung$^{1}$,  Hazem Youssef$^{2}$, and Moritz Roidl$^{2}$ % <-this % stops a space
\thanks{This work was partially supported by Bundesministerium für verkehr und digitale infrastruktur (BMVI) in the context of the project 45KI02B021 "Silicon Economy Logistics Ecosystem"}% <-this % stops a space
\thanks{$^{1}$Faculty of Electrical Engineering and Information Technology, Technische Universität Dortmund, 44227 Dortmund, Germany}%
\thanks{$^{2}$Chair of Material Handling and Warehousing, Faculty of Mechanical Engineering,
        Technische Universität Dortmund, 44227 Dortmund, Germany }
\thanks{{\tt \footnotesize \{yiu-ming.chung, hazem.youssef, moritz.roidl\} @tu-dortmund.de}}%
}

\begin{document}

\maketitle
\thispagestyle{empty}
\pagestyle{empty}

%%%%%%%%%%%%%%%%%%%%%%%%%%%%%%%%%%%%%%%%%%%%%%%%%%%%%%%%%%%%%%%%%%%%%%%%%%%%%%%%
\begin{abstract}

% This paper presents a novel Distributed Timed Elastic Band (DTEB) Planner that combines Prioritized Planning with Timed Elastic Band Planner, for multi-robot trajectory planning in production facilities. The proposed planner is able to reactively avoid imminent collisions as well as predictively resolve potential deadlocks among the team of robots, while navigating in a complex environment. The results of our simulation demonstrate the reliable performance and the versatility of the planner in different environment settings.
Autonomous navigation of mobile robots is a well-studied problem in robotics. However, the navigation task becomes challenging when multi-robot systems have to cooperatively navigate dynamic environments with deadlock-prone layouts. %Many existing planners treat such systems in a decentralized manner with trajectories generated for individual robots only. Other approaches plan trajectories in an offline fashion which hinders real-time reactive performance.
We present a Distributed Timed Elastic Band (DTEB) Planner that combines Prioritized Planning with the online TEB trajectory Planner, in order to extend the capabilities of the latter to multi-robot systems. 
The proposed planner is able to reactively avoid imminent collisions as well as predictively resolve potential deadlocks among a team of robots, while navigating in a complex environment. 
%In addition, this work contributes an improved obstacle model, a novel H-signature for homotopy classes evaluation, and an open-sourced trajectory sampling approach based on Voronoi diagrams. 
The results of our simulation demonstrate the reliable performance and the versatility of the planner in different environment settings. The code and tests for our approach are available online.
\end{abstract}

%%%%%%%%%%%%%%%%%%%%%%%%%%%%%%%%%%%%%%%%%%%%%%%%%%%%%%%%%%%%%%%%%%%%%%%%%%%%%%%%
\section{INTRODUCTION}

% Modern production facilities are becoming more reliant on multiple mobile robots to perform material handling operations. To safely navigate in the environment, mobile robots should predict and avoid collisions from each other during their online trajectory generation. 
% In production facilities, there can be narrow corridors and crossroads which may induce potential deadlocks between robots. A competent trajectory planner should be able to reactively avoid imminent collisions, and predictively avoid deadlocks between robots' future trajectories as well. 
Distributed multi-robot systems are increasingly being deployed in many settings. Dynamic production facilities and modern warehouses are examples of settings that are becoming more reliant on multiple robots to perform material handling operations. However, such settings can heavily affect the collaborative navigation performance of a robot team since there can be narrow corridors and crossroads which may induce deadlock situations between the robots. A competent trajectory planner should be able to reactively avoid imminent collisions, and predictively avoid deadlocks between the robots' future trajectories as well.
\begin{figure}[hbt]
  \centering
    \begin{subfigure}[b]{0.22\textwidth}
        \includegraphics[width=0.95\linewidth]{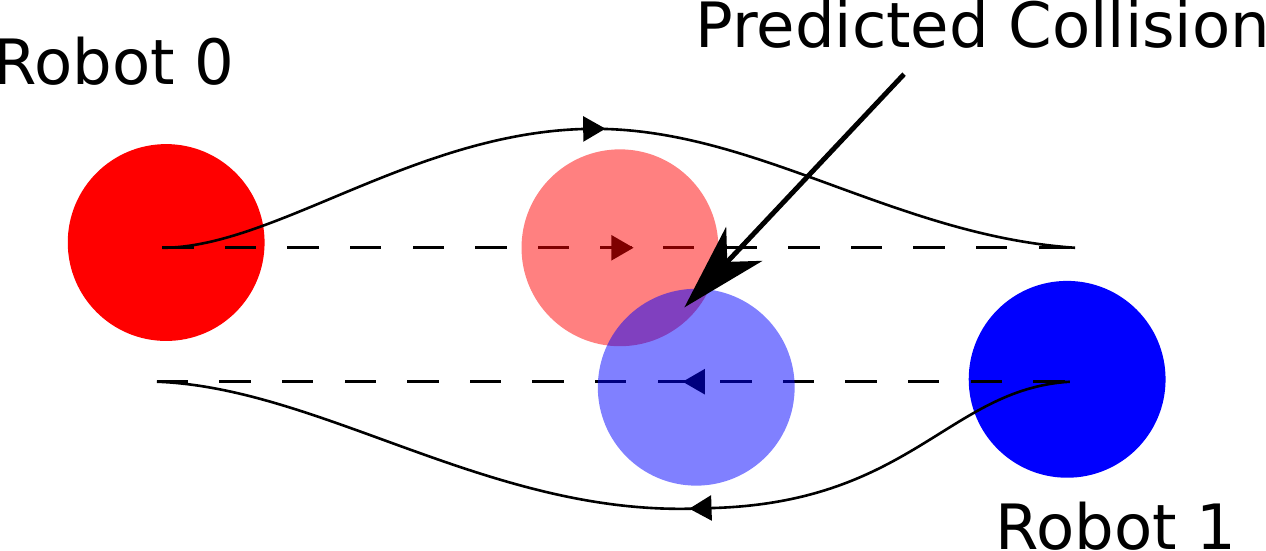}
        \subcaption{Reactive Avoidance}
        \squeezeup
        \label{fig:avoidance_a}
    \end{subfigure}
    \begin{subfigure}[b]{0.259\textwidth}
        \includegraphics[width=0.95\linewidth]{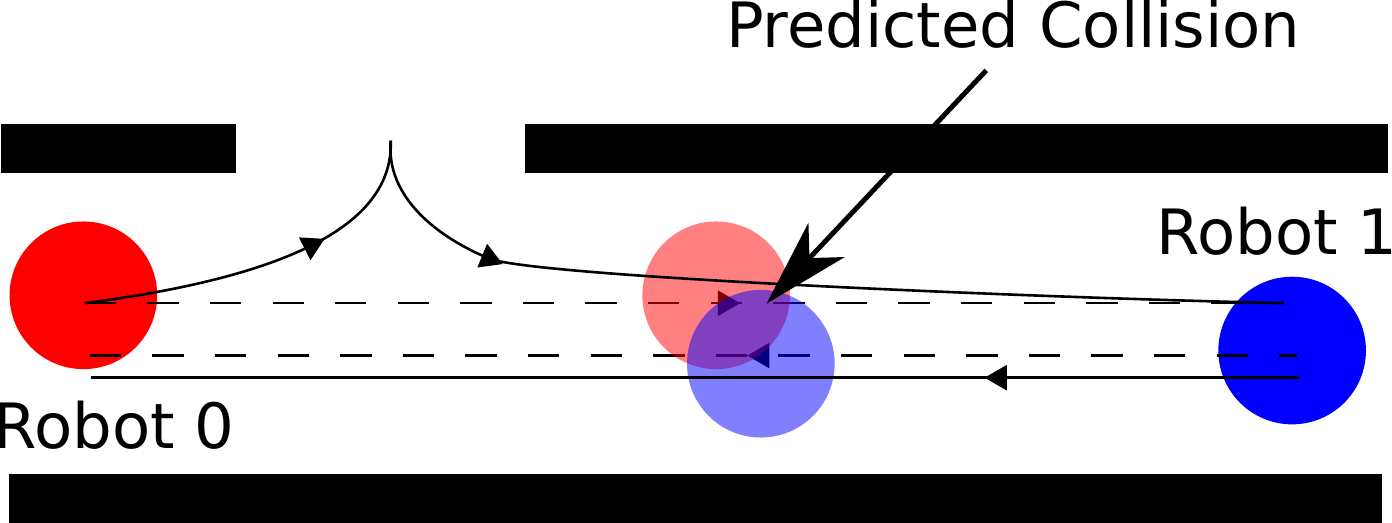}
        \subcaption{Predictive Avoidance}
        \squeezeup
        \label{fig:avoidance_b}
    \end{subfigure}
    \caption{Two types of collision avoidance. Dashed lines represent the intended trajectories. Solid lines represent the trajectories generated with collision avoidance.}
    \label{fig:avoidance}
\end{figure}
\par
We illustrate these two types of collision avoidance scenarios in \shortautoref{fig:avoidance}. In a reactive avoidance scenario, the robots only need to avoid each other at the position where a potential collision may occur based on their intended trajectories. In a predictive avoidance scenario, the avoidance maneuver needs to be performed in advance, at a position prior to where a potential collision may occur. The related works in online trajectory planning can handle reactive collision avoidance, but not predictive deadlock avoidance which is currently solved in general using discrete graph-based methods \cite{Hoy2014}. 
%in an offline manner (not really)
Therefore, in this paper, we propose a novel planner that combines Prioritized Planning with a Distributed Timed Elastic Band (DTEB) Planner to handle both reactive and predictive collision avoidance.

\section{Related Work}
\label{sec:related_work}

For the multi-robot trajectory planning problem, a classical approach is Velocity Obstacle (VO). VO considers the set of velocities of the ego robot that will result in a collision with another robot based on their current positions and velocities, and then finds the optimal control velocity that lies outside this forbidden velocity zone \cite{Fiorini1993}. There are many extensions to improve the performance of VO, such as Reciprocal Velocity Obstacle (RVO) \cite{vandenBerg2008} and Hybrid Reciprocal Velocity Obstacle (HRVO) \cite{Snape2011}. These papers have shown the effectiveness of VO-based approaches in local reactive collision avoidance. However, VO-based approaches optimize a single constant velocity input for a short period of time, which makes them unable to predictively avoid future deadlocks. 

%Another reactive planning approach is Buffered Voronoi Cell (BVC) method\mbox{\cite{BVC}}. BVC method uses robot positions to construct a Voronoi diagram which separates the robots into cells, and the borders of the cells retract by a safety distance to account for robot size. BVC method has theoretical guarantee of collision avoidance at next time step, however, it still has no consideration of future deadlocks outside BVCs.
\par
There are more advanced approaches which solve an Optimal Control Problem (OCP) and optimize a control sequence instead of a single control input. Recent examples include\cite{Defoort2010,Filho2017,Luis_2020,tordesillas2020mader}, %These approaches optimize a parameterized B-spline.\cite{Defoort2010,Filho2017,Luis_2020} enforce constraints on sample points of the B-spline, while\cite{tordesillas2020mader} uses the convex hull property to enforce constraints over the entire B-spline and uses their MINVO basis to obtain smaller convex hull to avoid conservatism. 
but these approaches have two issues that undermine their ability in predictive deadlocks avoidance. 

First, the prediction horizon is short for non-holonomic robots. For UAVs or holonomic robots, the trajectory can be planned efficiently using parametric B-spline as in \cite{Luis_2020,tordesillas2020mader}. For non-holonomic robots, the optimization problem is non-linear as in \cite{Defoort2010,Filho2017}, and is difficult to be efficiently solved. Therefore, these approaches limit the prediction horizon to a short period of time to ensure efficiency. However, the short prediction horizon reduces the ability to avoid deadlocks, as the collision avoidance constraint is no longer considered beyond the prediction horizon. In contrast, our approach solves an unconstrained least squares problem, using the soft-constraints proposed by the TEB approach in \cite{roesmann2017}, which has better computational efficiency than the original OPC \cite{roesmann2020online}. The improved efficiency allows a longer prediction horizon.

Second, the sequence of planning is unordered. However, the researches in Prioritized Planning (PP) show that the order of planning in a decoupled planning has a significant influence on the optimality and the completeness of the overall plan \cite{Parker2009} \cite{priority2001}. 
%Some deadlocks can only be solved if the decoupled planning is performed in a correct order. 
In particular, \cite{priority2001} shows that the corridor scenario in \shortautoref{fig:avoidance_b} can be solved if the robots plan their trajectories in this order: Robot 1 should plan first, and Robot 0 should avoid it at the doorway. If Robot 0 plans first, there is no solution for Robot 1, as it will collide with Robot 0 before reaching the doorway. 

% for multi-robot trajectory planning in production facilities. 
To address the above issues, this paper proposes a novel DTEB planner for multi-robot trajectory planning by contributing the following:
\begin{itemize}
    \item A Prioritized Planning scheme for reactive, as well as, predictive collision avoidance
    \item\hl{Extending the efficient TEB approach to include }an improved obstacle model that can handle robots moving at arbitrary trajectories.
    \item A \emph{novel} H-signature for the evaluation of homotopy
    classes of the generated trajectories. The evaluation is
    computationally efficient and able to handle arbitrary
    trajectories
    \item An open-source implementation of the Voronoi-diagram-based approach for trajectory sampling in narrow environments 
\end{itemize}

% The proposed planner is suitable for distributed multi-robot  systems  operating  in  dynamic  environments and can solve deadlock scenarios that are difficult to the previous methods.

% This paper contributes to the research area of multi-robot trajectory planning by: (1) proposing a novel decentralized multi-robot trajectory planner, which can handle scenarios that are difficult for the previous planners, (2) The proposed novel multi-robot trajectory planner is available in open source C++ code and integrated into Robot Operating System (ROS).

\section{Distributed Timed Elastic Band Planner}

The core idea of this paper is a distributed variant of the Timed Elastic Band (TEB) approach. Each robot in the multi-robot system cyclically broadcasts their latest planned trajectory, and the TEB approach combined with Prioritized Planning (PP) is used to plan its local optimal trajectory with consideration of the latest trajectory information. 

TEB approach \cite{roesmann2013} is a single-robot trajectory planning method and can be considered as a special form of \hl{open-loop} MPC. Instead of optimizing the control inputs, TEB approach directly optimizes the robot poses (and the time sequence) with soft-constraints and then extract control inputs from the poses \cite{roesmann2020online}. \par
Prioritized Planning (PP) is an efficient decoupled planning method to solve potential deadlocks between robots \cite{Parker2009}. In PP, each robot is given a priority. High priority robots plan first, while low priority robots plan later and treat high priority robots as dynamic obstacles. Previous works in PP use discrete graph-based methods, such as RRT with A* Algorithm \cite{Dewangan2017}.

Since the original TEB approach is designed for single robot but not the aforementioned purpose, we modify the original TEB approach and also incorporate new features as explained in the following subsections.

\begin{myhl}
\shortautoref{sec:assumptions} states the assumptions in the implementation.
\shortautoref{sec:optimal_control} formulates the planning problem as a soft-contrained optimization. \shortautoref{sec:obstacle_constraint} explain how collision avoidance between robot trajectories is included into the optimization.
\shortautoref{sec:trajectory_sharing}
explains the sharing of this trajectory information between robots.
\shortautoref{sec:H_sign} and \shortautoref{sec:voronoi} explain the improvement in homotopy class planning which is used to avoid local minimum problem in the optimization. \shortautoref{sec:PP} explains the Prioritized Planning scheme used to resolve deadlocks between trajectories.

\subsection{Assumptions}
\label{sec:assumptions}
\begin{enumerate}
\item The robot team can be non-holonomic and heterogeneous.
\item Each robot is controlled by an on-board computer and can communicate with each other within the planning horizon to share information as per \shortautoref{sec:trajectory_sharing}. 
\item The planning is asynchronous among the team but has a fixed frequency of 20Hz.
\end{enumerate}
\end{myhl}

\subsection{Optimal Control Formulation}
\label{sec:optimal_control}
For the proposed DTEB Planner, we adopted the existing optimal control formulation from the single-robot TEB Planner in \cite{roesmann2013} and \cite{roesmann2017} with a modification to the minimum obstacle separation constraint.
\par
A Timed-Elastic-Band $\mathcal{B}$ is a tuple of sequences of $n$ robot configurations \hl{($s_k = [x_k,y_k,\theta _k]^T)_{k=1,2,...,n}$} and time differences between them $(\Delta T_k)_{k=1,2,...,n-1}$:
\begin{equation}
   \mathcal{B} := \{ s_1, \Delta T_1, s_2, \Delta T_2 ,..., s_{n-1}, \Delta T_{n-1}, s_n \} 
   \label{eqn:teb}
\end{equation}
Both the configurations and time differences are optimized against a soft constrained time optimal cost function to obtain an optimized TEB $\mathcal{B}^*$ as the robot trajectory:
\begin{equation}
\mathcal{B}^{*} =\argminA_{\mathcal{B} \setminus \{ s_1,s_n \}} \tilde{V}(\mathcal{B})
\end{equation}
The time optimal cost function is given by:
\begin{equation}
\resizebox{0.91\hsize}{!}{
$\begin{aligned}
\tilde{V}(\mathcal{B}) = &\sum_{k=1}^{n-1} [\Delta T_k^2 + \phi(\mathbf{h}_k, \sigma_h) + \chi(\tilde{r}_k,\sigma_r) + \chi(\boldsymbol{\nu}_k,\sigma_\nu) \\ & +\chi(\mathbf{o}_k,\sigma_\mathbf{o}) + \chi(\boldsymbol{\alpha}_k,\sigma_\alpha)] + \chi(\boldsymbol{\alpha}_n,\sigma_\alpha)
\end{aligned}
$
}
\label{eqn:cost_function}
\end{equation}

\begin{myhl}
$\phi$ and $\chi$ are quadratic penalties approximating the kinematics, dynamics and collision-free constraints:
\begin{equation}
\phi(\cdot, \sigma) =  \sigma \norm{\cdot}^2_2.
\end{equation}
\begin{equation}
\chi(\cdot, \sigma) = \sigma \norm{ min \{ \mathbf{0}, \cdot \} }^2_2.
\end{equation}
where $\sigma$ is the scalar weight of the penalty term.

$\phi(\mathbf{h}_k, \sigma_h)$ is the penalty approximating the equality constraint of non-holonomic kinematics:
\begin{equation}
\mathbf{h}_k = 
\begin{pmatrix}
cos(\theta_k) + cos(\theta_{k+1}) \\
sin(\theta_k) + sin(\theta_{k+1}) \\
0
\end{pmatrix}  
\times 
\begin{pmatrix}
x_{k+1} - x_k\\
y_{k+1} - y_k\\
0 
\end{pmatrix}  
= 0
\end{equation}
and $\chi(\tilde{r}_k,\sigma_r)$, $\chi(\boldsymbol{\nu}_k,\sigma_\nu)$, $\chi(\boldsymbol{\alpha}_k,\sigma_\alpha)$ and $\chi(\mathbf{o}_k,\sigma_\mathbf{o})$ are the penalties approximating the inequality constraints of minimum turning radius, maximum velocities, maximum accelerations and minimum obstacle separation, respectively:
\begin{equation}
\tilde{r}_k = r_k - r_{min} \geq 0
\end{equation}
\begin{equation}
 \boldsymbol{\nu}_k = [{v}_{max} - |v_k|, {\omega}_{max} - |\omega_k|]^T \geq \mathbf{0}
\end{equation}
\begin{equation}
 \boldsymbol{\alpha}_k = [\mathrm{{a}}_{max} - |\mathrm{a}_k|, {\alpha}_{max} - |\alpha_k|]^T \geq \mathbf{0}
\end{equation}
\end{myhl}
\hl{The above constraints are the same as the original TEB approach (for details, see\cite{roesmann2017}), while the inequality constraint of minimum obstacle separation is modified:}
\begin{equation}
\mathbf{o}_k(s_k) =
\begin{pmatrix}
\delta(s_k,O_1(T_k))\\
\delta(s_k,O_2(T_k))\\
\vdots\\
\delta(s_k,O_R(T_k))
\end{pmatrix}  
-
\begin{pmatrix}
\delta_{min}\\
\delta_{min}\\
\vdots\\
\delta_{min}
\end{pmatrix}
\geq \mathbf{0}
\end{equation}
where $\delta(s_k,O(T_k))$ describes the minimal Euclidean distance between obstacle $O$ at time $T_k$ and pose $s_k$ of the ego robot.
\par
We modify the $\mathbf{o}_k(s_k)$ function because the original TEB Planner has the limitation that it only considers dynamic obstacles moving in constant velocity. Such limitation is not suitable for our application, since other robots, considered as obstacles by the ego robot, move in arbitrary trajectories. 
% We extend the TEB Planner by introducing a new model of obstacles moving in arbitrary trajectories
\par
 The corresponding functions for the obstacle separation $\delta$ and homotopy class evaluation $\mathcal{H}$ are then extended accordingly as discussed in \shortautoref{sec:obstacle_constraint} and \shortautoref{sec:H_sign}, respectively.
\subsection{Obstacle Separation Extension}
\label{sec:obstacle_constraint}

The dynamic obstacle separation $\delta$ is calculated using obstacle trajectory. A discrete trajectory $\tau$ can be directly obtained from TEB $\mathcal{B}$ in \shortautoref{eqn:teb} by summing the total time up to that pose. 
\begin{equation}
\tau:= \{(T_1, s_1),(T_2, s_2),...,(T_n, s_n)\}
\end{equation}
where $T_k = \Delta T_1+\Delta T_2+...+\Delta T_{k-1}$
\par
To estimate the obstacle pose at any time $t$, the obstacle pose is interpolated between two successive points in its trajectory. For holonomic robot, the obstacle pose is estimated by linear interpolation. For non-holonomic robot, each segment is assumed to be a circular arc.
\par
\begin{figure}[htb]
  \centering
    \includegraphics[width=0.9\linewidth]{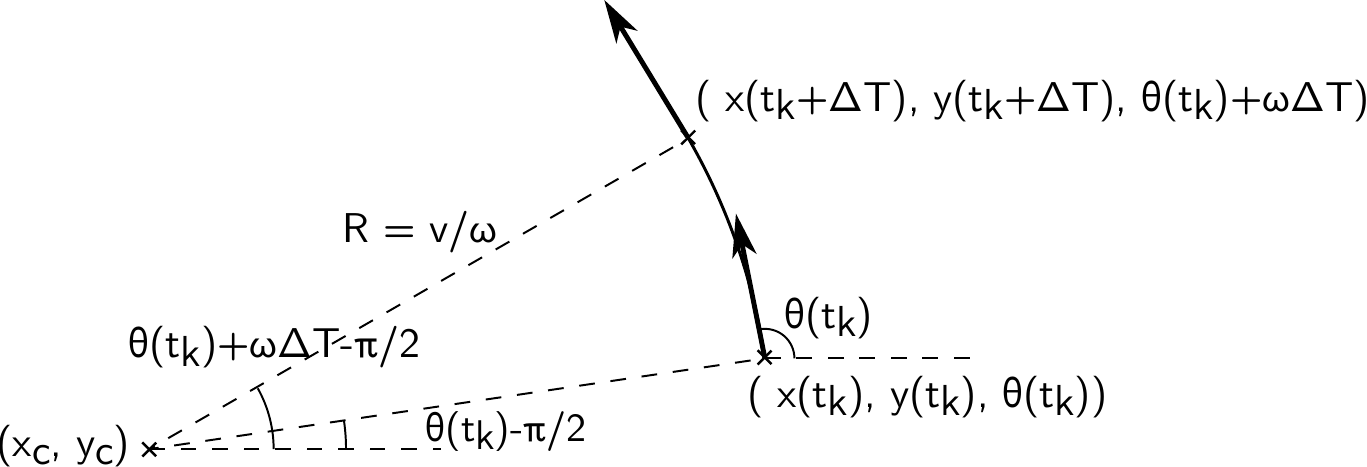}
    \squeezeup
    \caption{Constant circular motion}
    \label{fig:constantArc}
    \squeezeup
\end{figure}

The implementation of linear interpolation of obstacle pose is trivial. 
%To interpolate the robot position from $t_k$ to $t_k + \Delta T$:
%\begin{equation}
%\begin{cases}
%& x(t_k + \Delta T) = x(t_k) + v_x(t_k)  \Delta T \\
%& y(t_k + \Delta T) = y(t_k) + v_y(t_k)  \Delta T \\
%& \theta(t_k + \Delta T) = \theta(t_k) + \omega(t_k)  \Delta T \\
%\end{cases} 
%\label{eqn:linearInterpolation}
%\end{equation}
The circular interpolation is computed by considering the geometry of the circular arc, as shown in Fig.{\ref{fig:constantArc}}.
\par 
By trigonometry:
\begin{equation}
 \resizebox{0.91\hsize}{!}{
$\begin{cases}
& x(t_k + \Delta T) = x(t_k) + \frac{v(t_k)}{\omega(t_k)}(sin( \theta(t_k) + \omega(t_k) \Delta T) - sin(\theta(t_k)) ) \\
& y(t_k + \Delta T) = y(t_k) - \frac{v(t_k)}{\omega(t_k)}(cos( \theta(t_k) + \omega(t_k) \Delta T) - cos(\theta(t_k)) ) \\
& \theta(t_k + \Delta T) = \theta(t_k) + \omega(t_k)  \Delta T
\end{cases}$
}
\label{eqn:circularInterpolation}
\end{equation}
\par
After obtaining the obstacle pose, the obstacle footprint is projected onto the pose, and the separation between the obstacle and the ego robot $\delta(s_k,O)$ can be computed accordingly by solving the geometric problem of minimum separation between two 2D shapes. 
\par

The above calculations require linear and angular velocities. These velocities need to be computed during trajectory generation, and then the trajectories containing poses, velocities and time are published among the robots. 

\subsection{Trajectory Sharing}
\label{sec:trajectory_sharing}

The latest planned trajectory is shared among robots through unidirectional broadcasting. The actual implementation is in the form of standard ROS publisher-subscriber communication. The trajectory is defined by the following message:
\begin{lstlisting}
std_msgs/Header header
time priority
teb_local_planner/ObstacleFootprint footprint
teb_local_planner/TrajectoryPointSE2[] trajectory
\end{lstlisting}

$header$ which contains the timestamp of message generation is used to compensate communication delays while packet loss can be intrinsically handled by TCP. $priority$ is used in Prioritized Planning which will be discussed in \shortautoref{sec:PP}. $footprint$ describes the dynamic obstacle (other robot) footprint,\hl{ which can be point, circle, line, pill shape and polygon.}  $trajectory$ is represented by a series of trajectory points. Each point contains the $SE(2)$ pose [x, y, $\theta$], time and velocity which are necessary for the calculations in \shortautoref{sec:obstacle_constraint}.

\subsection{New H-signature for Homotopic Planning}
\label{sec:H_sign}

Parallel planning of multiple trajectories in different homotopy classes is a feature in TEB planner to avoid the optimization of TEB being trapped in a local minimum {\cite{roesmann2015}}. \hl{TEB approach uses H-signature to distinguish homotopy classes. The detailed definition of H-signature is given in [15]. The idea is briefly explained as follows:}
\par
\begin{myhl}
The H-signature for a robot trajectory $\tau$ is the line integral of the magnetic field $\mathbf{B}$ created by a virtual current flowing along the trajectory of the obstacle(s):
\begin{equation}
\mathcal{H}(\tau) = \int _{\tau} \mathbf{B}(l) \cdot d \mathbf{l}
\label{eqn:Hsigature}
\end{equation}
where the magnetic field $\mathbf{B}$ at any arbitrary point $\mathbf{r}$, due to the current flow in the obstacle trajectory $\mathcal{S}$, is given by: 
\begin{equation}
\mathbf{B}(\mathbf{r}) = \frac{1}{4\pi} \int _{\mathcal{S}} \frac{(\mathbf{x}-\mathbf{r}) \times  d \mathbf{x} }{ \norm{\mathbf{x}-\mathbf{r}}^3 }
\label{eqn:bField}
\end{equation}
\end{myhl}
\par
\hl{Two trajectories generate the same H-signature if they are in the same homotopic class, according to Ampere's law: for a loop enclosing the obstacle trajectory, the line integral is one, otherwise zero. 
Considering two trajectories $\tau_1$ and $\tau_2$ which have the same start and goal and are in the same homotopy class, the loop $\tau_1 \cup -\tau_2$ will not enclose an obstacle, the H-signature of the loop will be zero:}
\begin{equation}
\resizebox{0.91\hsize}{!}{$
\begin{aligned}
& \mathcal{H}(\tau_1 \cup -\tau_2) = \int _{\tau_1 \cup -\tau_2} \mathbf{B} \cdot d \mathbf{l} = \int _{\tau_1} \mathbf{B} \cdot d \mathbf{l} - \int _{\tau_2} \mathbf{B} \cdot d \mathbf{l} = 0 
\\
\iff & \mathcal{H}(\tau_1) = \mathcal{H}(\tau_2)
\end{aligned}
$}
\label{same_H}
\end{equation}
\par
However, the calculation of the H-signature requires double integration. For the original TEB planner, this is not a problem, since it only considers obstacles with constant velocity. In that case, the obstacle trajectory $\mathcal{S}$ will be a straight line in the 3D space and the $\mathbf{B}$-field has an analytical solution \cite{Bhatt2011}. 
\par
In this work, we propose an alternative H-signature which requires only one line integral instead of two as proposed in the original paper. This is important as the H-signature needs to be efficiently computed multiple times in each planning cycle.
The proposed $\mathbf{B}$-field is a fictional magnetic field that at any point on the obstacle trajectory $\mathcal{S}$, the $\mathbf{B}$-field is considered as a local horizontal planar magnetic field generated by an upward current I.
\par
Let $\mathbf{x} = (f(t), g(t), t)^T$ be a point on the obstacle trajectory $\mathcal{S}$. At any point $\mathbf{r}$ in space-time $(x, y, t)^T$, the $\mathbf{B}$ field experienced by the robot is: \begin{equation}
\resizebox{0.91\hsize}{!}{$
\mathbf{B}(\mathbf{r}) = \frac{1}{2 \pi}(\frac{y-g(t)}{(x-f(t))^2+(y-g(t))^2}, -\frac{x-f(t)}{(x-f(t))^2+(y-g(t))^2}, 0)^T
$}
\label{eqn:newB}
\end{equation}
Or in cross product form:
\begin{equation}
\mathbf{B}(\mathbf{r}) = \frac{1}{2 \pi} \frac{(\mathbf{x}-\mathbf{r}) \times \uvec{k} }{ \norm{\mathbf{x}-\mathbf{r}}^2 }
\end{equation}
The factor $\frac{1}{2\pi}$ is chosen for scaling and has no actual effect.\\
This $\mathbf{B}$-field shows correct behaviours in distinguishing homotopy classes during testing.
\par
After analysing this $\mathbf{B}$-field, we find that this $\mathbf{B}$-field is a good approximation under general conditions.\\ 
The curl of field $\mathbf{B}$, $\nabla \times \mathbf{B}$ is:
\begin{equation}
\resizebox{0.91\hsize}{!}{$
\begin{aligned}
& \frac{f'(t) ((x-f(t))^2 - (y-g(t))^2)+2g'(t)(x-f(t))(y-g(t))}{((x - f(t))^2 + (y - g(t))^2)^2} \uvec{\imath} \\
 & + \frac{g'(t) ((y-g(t))^2 - (x-f(t))^2)+2f'(t)(x-f(t))(y-g(t))}{((x - f(t))^2 + (y - g(t))^2)^2} \uvec{\jmath}\\
 & + 0 \uvec{k}\\
\end{aligned}
$}
\label{eqn:curl}
\end{equation}

\begin{figure}[htb]
  \centering
    \includegraphics[width=0.5\linewidth]{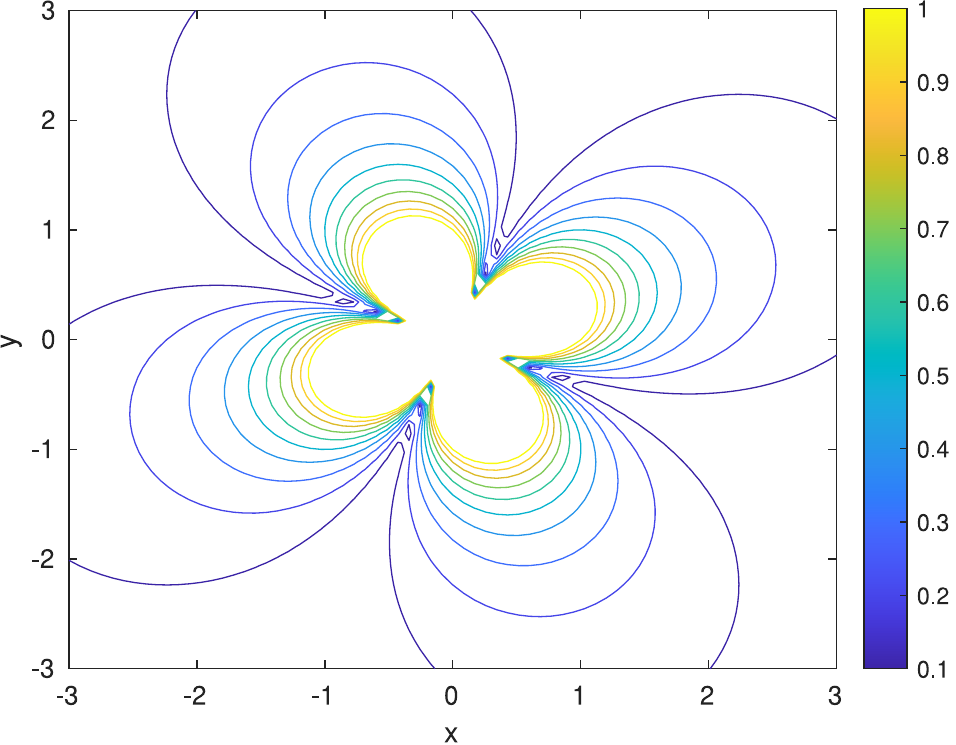}
    \squeezeup
    \caption{Contour plot of $|\frac{x^2 - y^2 +2xy}{(x^2+y^2)^2}|$}
    \label{fig:contour_plot}
    \squeezeup
\end{figure}

For the x and y component, the numerator part is a hyperbolic paraboloid with the position of the obstacle at the centre. The denominator part is the inverse of obstacle distance to the fourth power. For example, if the obstacle velocity $f'(t) = g'(t) = 1$ with obstacle position (f(t), g(t)) being the origin, the x and y components become $\frac{\pm (x^2-y^2)+2xy}{(x^2+y^2)^2}$, which equates to zero other than at the position of the obstacle, as shown in \shortautoref{fig:contour_plot} .

By Kelvin–Stokes theorem, the line integral along a loop is equal to the surface integral of the curl over the surface bounded by the loop. The surface integral will be almost zero when the loop does not enclose the obstacle and does not have a small obstacle separation compared to obstacle velocity. For two trajectories that have the same homotopic class, the overall H-signature is approximately zero and fulfil \shortautoref{same_H}. Therefore, the proposed alternative H-signature satisfies our application.

\subsection{Voronoi Roadmap}
\label{sec:voronoi}
Since our DTEB Planner is designed to be used in industrial environment, we need to further improve the planner's ability in sampling trajectories for homotopy class planning. 
\par
\begin{figure}[!htb]
  \centering
    \begin{subfigure}[b]{0.249\textwidth}
        \includegraphics[width=0.90\linewidth]{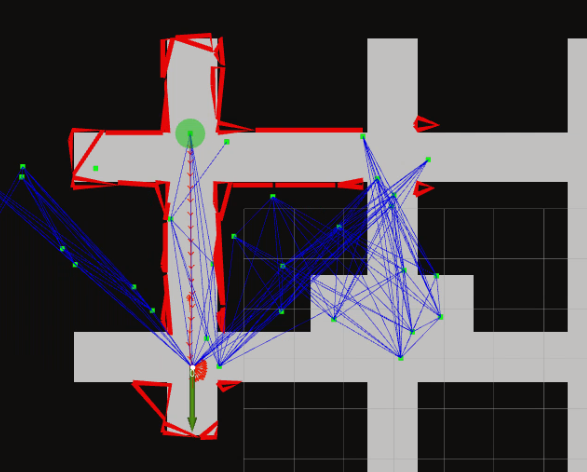}
        \captionsetup{width=.9\linewidth}
        \subcaption{Original Probabilistc Roadmap}
        \squeezeup
        \label{fig:roadmap_a}
    \end{subfigure}
    \begin{subfigure}[b]{0.229\textwidth}
        \includegraphics[width=0.90\linewidth]{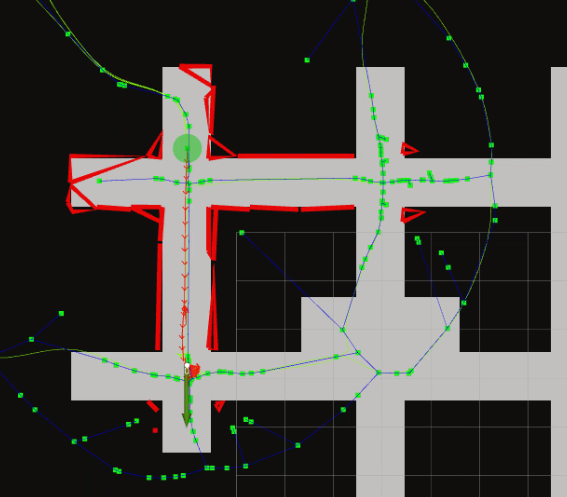}
        \captionsetup{width=.9\linewidth}
        \subcaption{Proposed Voronoi Roadmap}
        \squeezeup
        \label{fig:roadmap_b}
    \end{subfigure}
    \caption{Homotopy class planning using roadmap}
    \label{fig:voronoi_rviz}
\end{figure}

The original TEB planner creates a Probabilistic Roadmap in each planning cycle, and candidate trajectories are sampled from the roadmap by comparing their homotopy classes with the existing candidates \cite{roesmann2015}. The roadmap is efficient in an open space scenario where the obstacle space constitutes only a small part of the overall workspace. However, in a warehouse environment where there are only narrow aisles in which the robots can travel, a simple probabilistic roadmap can hardly sample any feasible trajectory as shown in \shortautoref{fig:roadmap_a}.
\par
We extend the homotopy class planning by using a generalized Voronoi diagram (GVD). Similar approaches of homotopy class planning by sampling paths from GVD are proposed in \cite{Kuderer2014} and \cite{Roesmann2017IntegratedOT}. In \cite{Kuderer2014} a grid-based approach is used, which has a total computation time of GVD construction and graph search in the magnitude of 100ms. Such computation time is too large for TEB planner, as the planner is expected to run at 20Hz. \cite{Roesmann2017IntegratedOT} reports an implementation using Boost library which is based on sweepline algorithm in \cite{Fortune1987} and has computation time in the magnitude of 10ms. However, the actual implementation is not explained in the paper or open-sourced. Therefore, we implement a new open-source ROS plugin using OpenCV and Boost library. The total time for GVD and graph search takes only 10-20ms for a 300x300 pixels local costmap.

\subsection{Prioritized Planning}
\label{sec:PP}

The researches in Prioritized Planning show that some deadlocks can only be solved by Prioritized Planning, but not Reciprocal Avoidance \cite{CAP2015} \cite{priority2001}. In this work, we propose a 2-stage priority scheme that takes the advantages of both Reciprocal Avoidance and Prioritized Planning. 
\par
The planning is divided into 2 stages according to the robot's Euclidean distance to the obstacle. The first stage is the prioritized stage. In this stage, Prioritized Planning similar to {\cite{CAP2015}} is carried out, with the difference that the planning order of our planner is not synchronized in sequential manner. The robots are given an initial priority. The lower priority robots treat the higher priority robots as moving obstacles following their latest published trajectories. This priority can be manually assigned or determined by heuristic metrics. In our implementation, for simplicity, the priority is determined according to chronological order. The robot with a more recent navigation task has a higher priority. The planner first tries to find a locally optimal solution according to the initial priority. If there is no feasible solution, the planner would raise the priority of the ego robot, similar to the method used in \cite{andreychuk2018techniques}.
\par
The second stage is reciprocal avoidance stage, in this stage the ego robot considers all robots as obstacles regardless of their priority. The target of this stage is to find a solution that can only be reached by a reciprocal strategy. Since all other robots are considered as moving obstacles by the ego robot, this stage also ensures safety when the robots are in the vicinity.

\section{Experiments}
In this section, we demonstrate the effectiveness of our planner in comparison to the related work in terms of the reactive and predictive avoidance scenarios. Two separate sets of tests are designed for the two scenarios respectively. The first test is designed for reactive avoidance, in which the robots on the perimeter of a circle have to travel to their antipodal positions. This circle test is also widely used in other multi-robot planners, such as \cite{Snape2011} \cite{Defoort2010}. The second test is designed for predictive avoidance, in which the robots  have to travel in a warehouse map consisting of narrow corridors, so as to simulate an environment full of potential deadlocks. 

\subsection{Reactive Avoidance Tests}   

\begin{figure}[htb]
  \centering
    \includegraphics[width=0.45\linewidth]{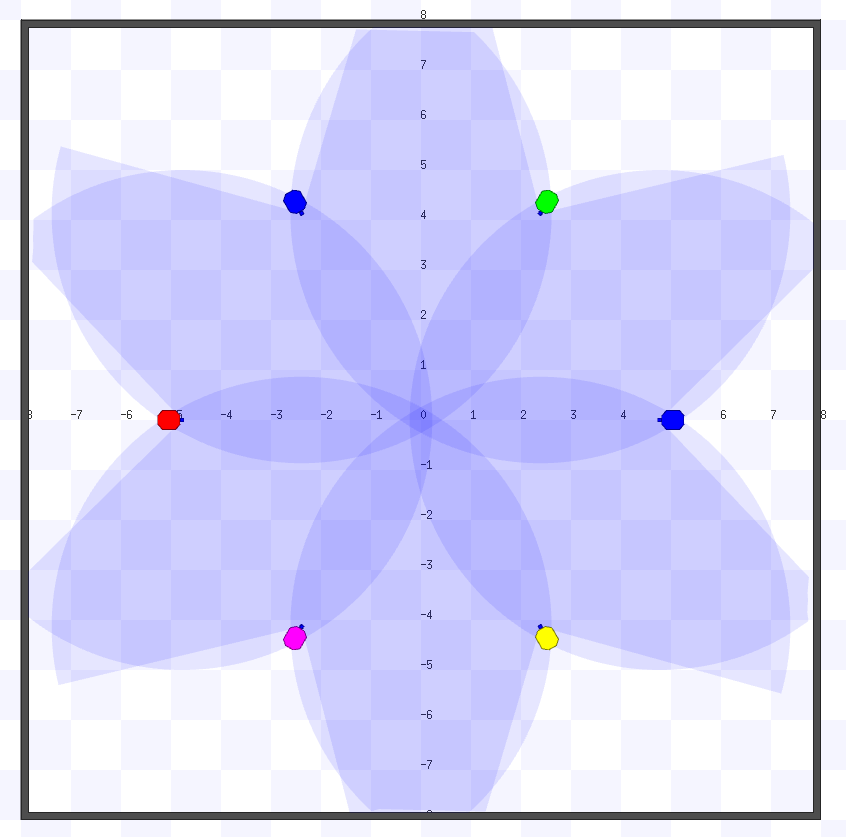}
    \squeezeup
    \caption{Setup for testing of reactive behaviour between six robots}
    \squeezeup
    \label{fig:stage_six}
\end{figure}
\par
To study the reactive behaviour involving multiple robots, the planner is tested in Stage simulator using the setup shown in \shortautoref{fig:stage_six}. Six robots are placed in a circle of diameter of 10 m, and the antipodal positions are used as the goals. Each test is repeated five times to ensure consistent results. The computer environment is Ubuntu 20.04 with an Intel(R) Xeon(R) X5675 @3.07GHz CPU. The simulation is run at 60\% speed due to computational burdens of running multiple instances of the planner. 

\begin{table}[!htb]
\caption{Comparison of the results of six-robot test between HRVO and the proposed DTEB Planner}
\label{tab:comparison_reactive}
\squeezeup
\resizebox{0.48\textwidth}{!}{%
\tiny
\begin{tabular}{lcc}
\hline
\multicolumn{1}{r}{} &
  \begin{tabular}[c]{@{}c@{}}DTEB with\\Prioritised Planning\end{tabular} &
  HRVO \\\hline
Total travel distance (m) &
  62.16 ±1.06&
  60.64 ±0.34\\
Time to finish (s) &
  12.86 ±0.53&
  14.92 ±1.22\\
Computation time per robot (ms) &
  10.30 ±9.45&
  3.94 ±0.70\\\hline
\end{tabular}
}
\end{table}
Since HRVO is a well-proven planner for reactive collision avoidance as mentioned in \shortautoref{sec:related_work}, the same test is performed with a HRVO planner for comparison. For this purpose, the ROS implementation by \cite{Claes2012} is used. \shortautoref{tab:comparison_reactive} shows the average values of the distance travelled, the duration to finish a run and the computation time, along with their standard deviations.

\begin{figure}[!htb]
  \centering
    \begin{subfigure}[b]{0.239\textwidth}
        \includegraphics[width=0.95\linewidth]{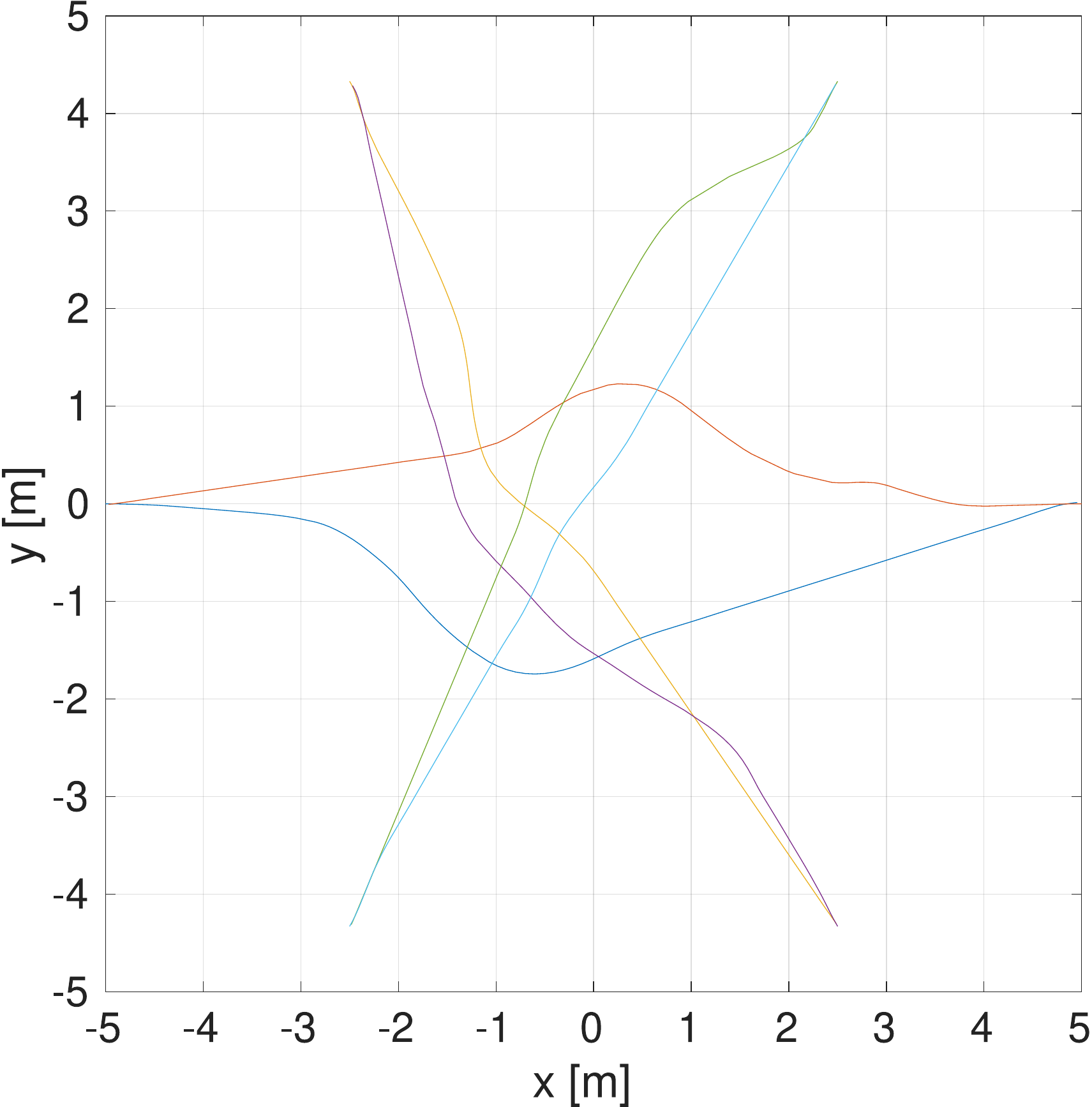}
      \captionsetup{width=.9\linewidth}
        \subcaption{Sample trajectory of six robots controlled by DTEB Planner in an open space}
        \squeezeup
        \label{fig:teb1}
    \end{subfigure}
    \begin{subfigure}[b]{0.239\textwidth}
        \includegraphics[width=0.95\linewidth]{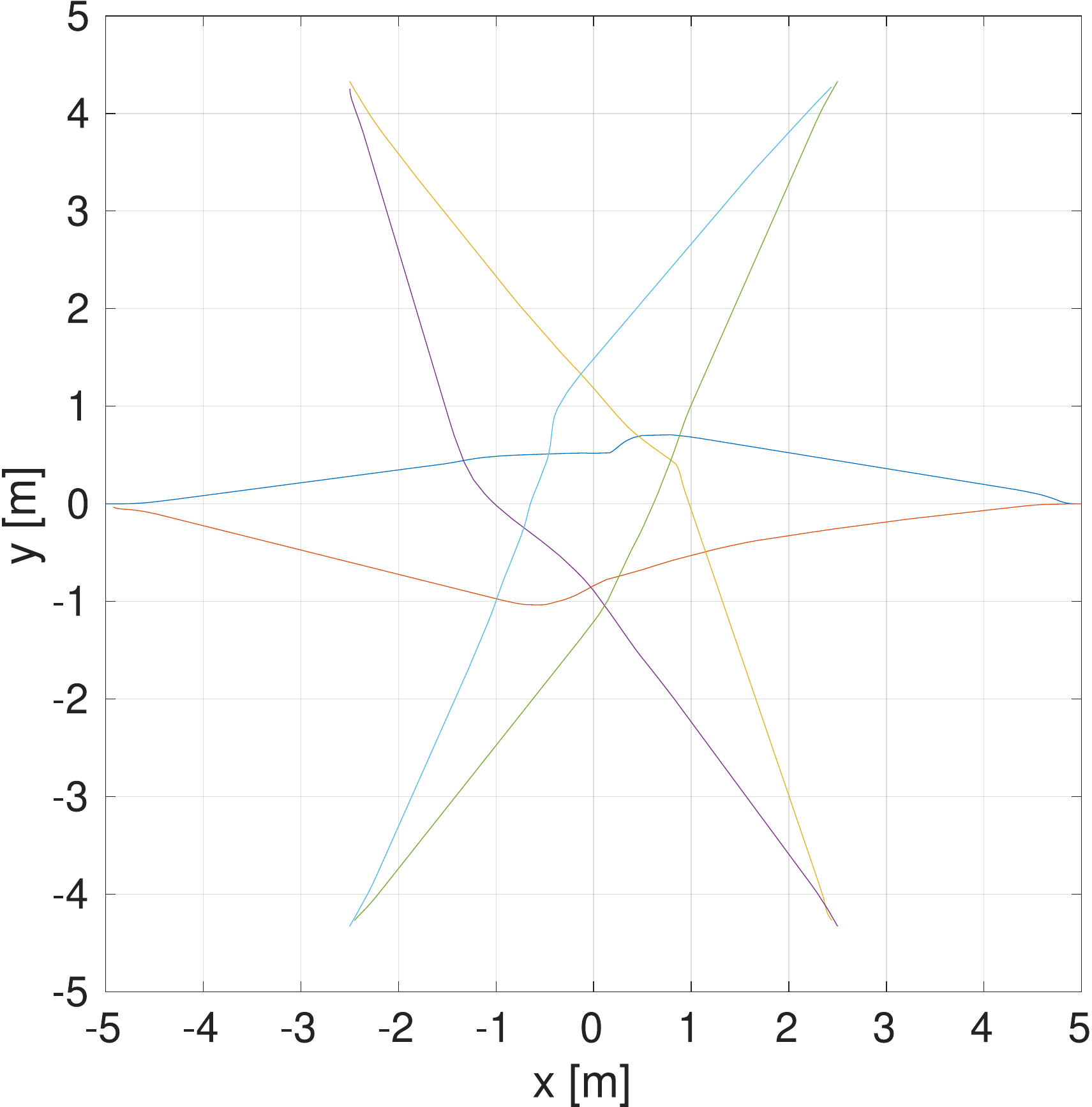}
        \captionsetup{width=.9\linewidth}
        \subcaption{Sample trajectory of six robots controlled by HRVO Planner in an open space}
        \squeezeup
        \label{fig:hrvo}
    \end{subfigure}
    \caption{}
    \squeezeup
    \label{fig:sample_trajectories}
\end{figure}

As seen in \shortautoref{fig:sample_trajectories}, \hl{the trajectories of DTEB are less regular than HRVO. It is caused by the symmetrical setting in this scenario as it has the same cost for a robot to go around another robot on left or on the right. In contrast, HRVO has a heuristic rule to favour one side and produce more regular trajectories (For details see\cite{Snape2011}).} Although the proposed DTEB Planner has less regular trajectories, it outperforms the HRVO planner in travel time by a small margin. The major reason is that the HRVO planner has a simple (non-time optimal) homing strategy for the selection of preferred velocity. The planner performs unnecessary deceleration and turning in place at the goal to satisfy the goal orientation. 
%In contrast, the DTEB planner takes a time optimal approach, in which the time optimal deceleration and turning are already considered as a part of the planning.
In contrast, the DTEB planner already considers a time optimal deceleration and turning as part of the planning.

\subsection{Predictive Avoidance Tests}
The existing HRVO planners do not consider deadlocks in the long-term future time. Therefore, we mainly use a Multi-Robot Route Planner $tuw\_multi\_robot$ in \cite{Binder2019} for comparison instead. The \hl{TUW} Route Planner is an offline trajectory planner, which generates a centralized routing plan consisting of robot paths from start to goal positions and their temporal information. The robots' local planner then simply follow the routing plan. 

\begin{figure}[htb]
  \centering
    \includegraphics[width=0.5\linewidth]{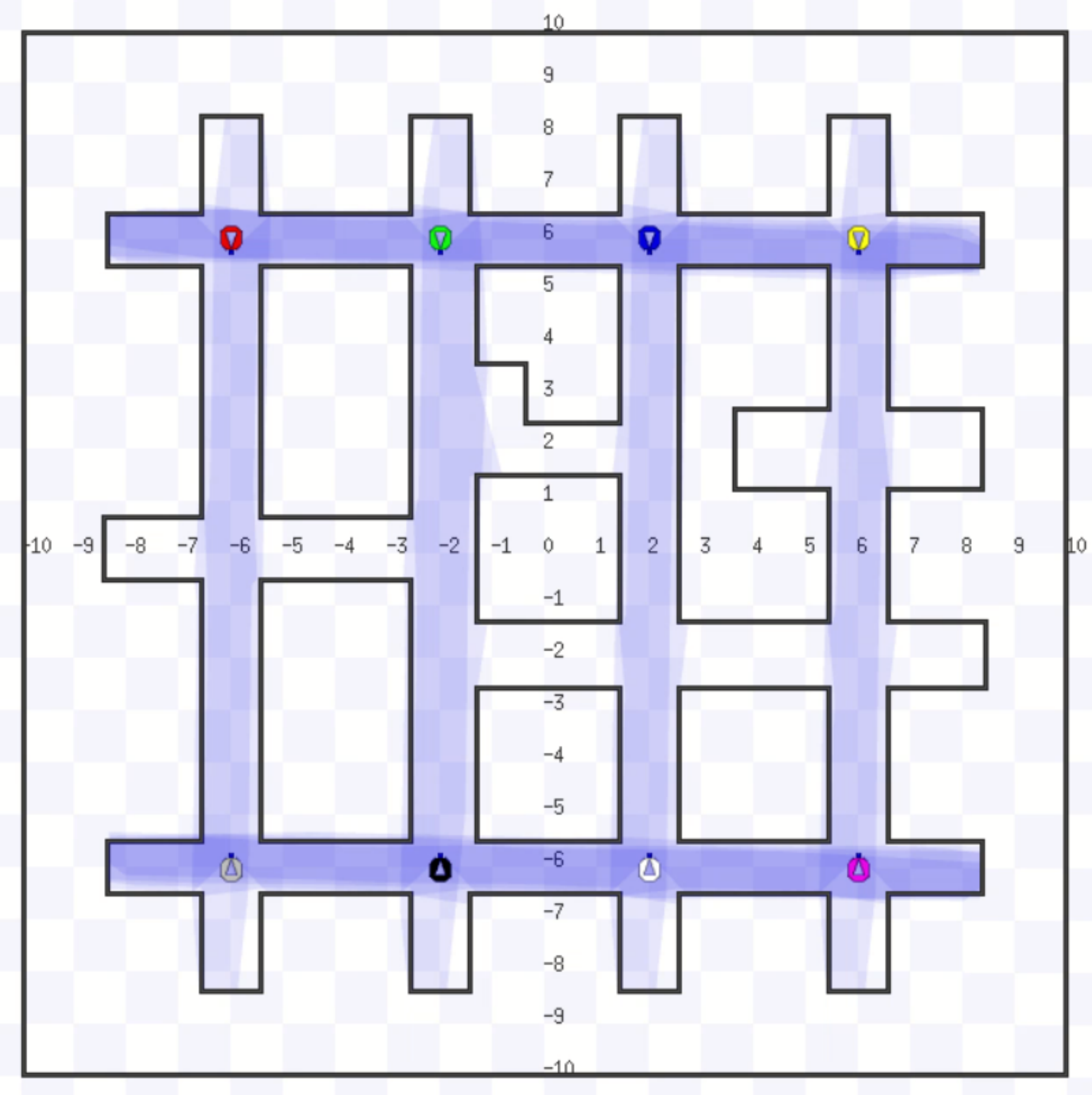}
    \squeezeup
    \caption{Eight-robot predictive avoidance test in a warehouse map}
    \label{fig:stage_8_robots}
    \squeezeup
    \end{figure}

To further examine the proposed planner's capability in predictive collision avoidance, we test the planner using more robots in Stage simulator. A warehouse-like map from \cite{Binder2019} is modified for the testing, as shown in \shortautoref{fig:stage_8_robots}.

\begin{figure}[htb]
  \centering
    \begin{subfigure}[b]{0.239\textwidth}
        \centering
        \includegraphics[width=0.95\linewidth]{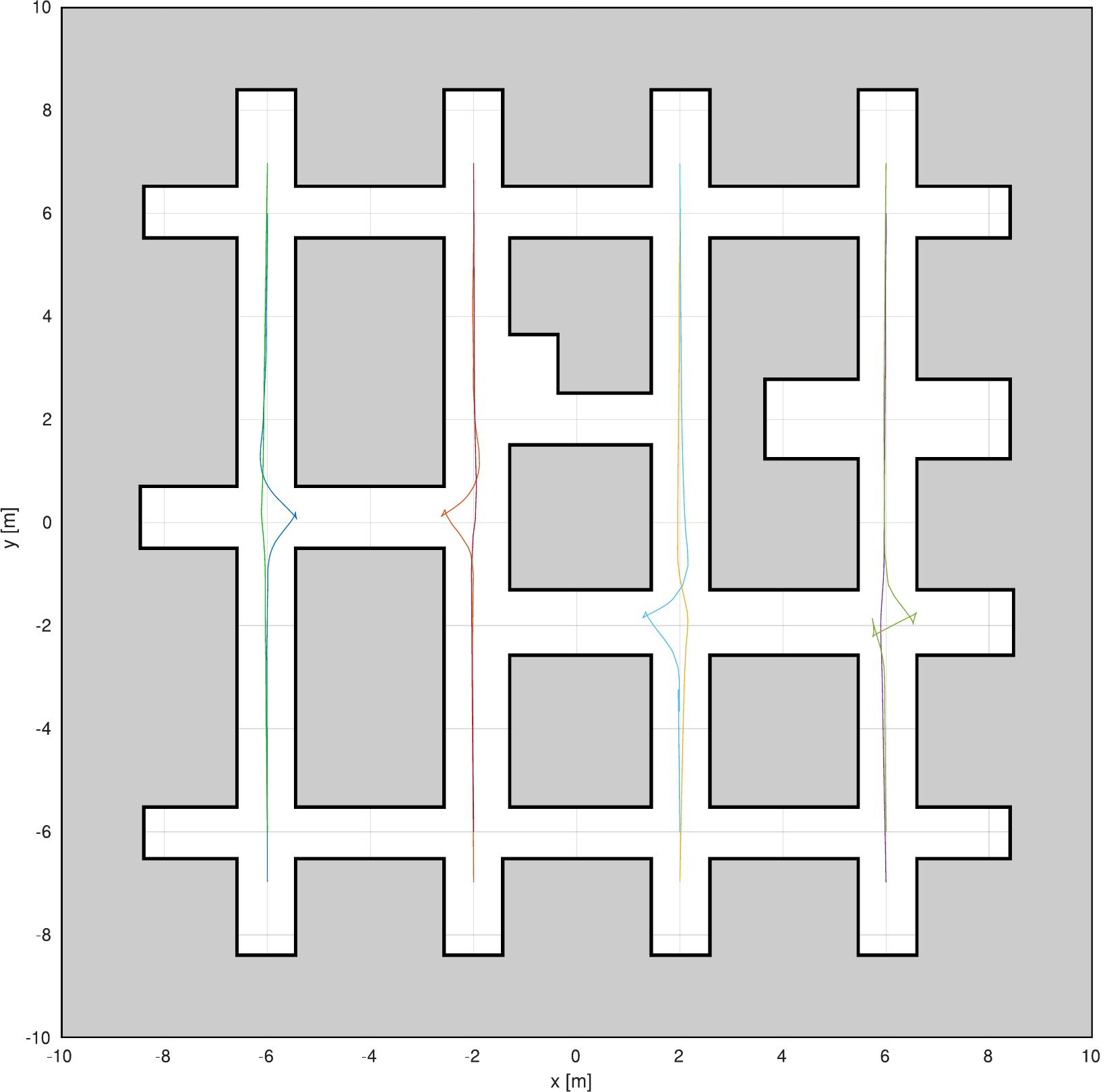}
        \captionsetup{width=.9\linewidth}
        \subcaption{Sample trajectory of eight robots controlled by DTEB Planner in the warehouse map}
        \squeezeup
        \label{fig:warehouse_teb_1}
    \end{subfigure}
    \begin{subfigure}[b]{0.239\textwidth}
        \centering
        \includegraphics[width=0.95\linewidth]{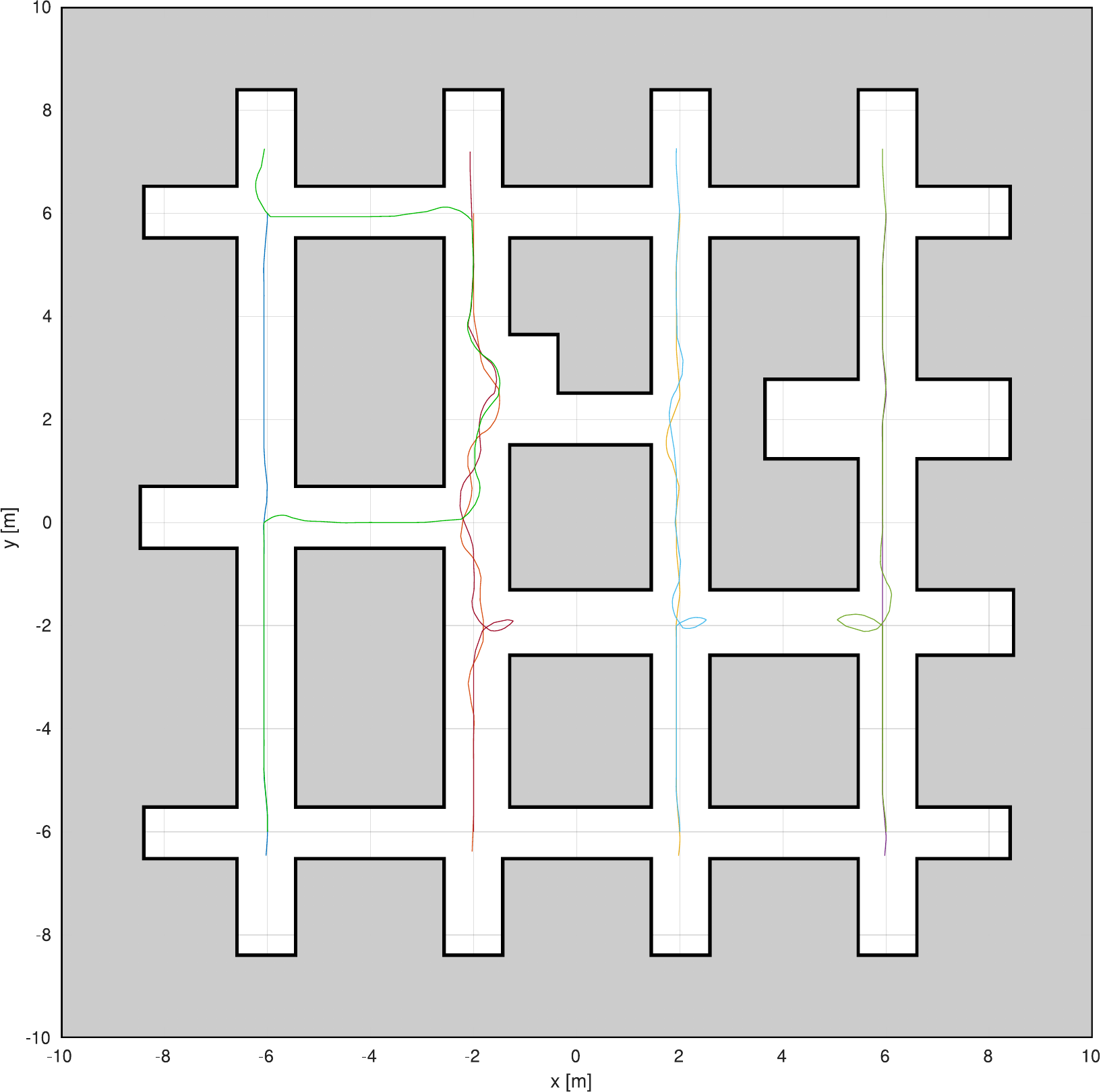}
        \captionsetup{width=.9\linewidth}
        \subcaption{Sample trajectory of eight robots controlled by TUW Route Planner in the warehouse map}
        \squeezeup
        \label{fig:warehouse_tuw_1}
    \end{subfigure}
    \caption{Predictive test using opposite positions as goals}
    \squeezeup
\end{figure}
\begin{figure}[htb]
  \centering
    \begin{subfigure}[b]{0.239\textwidth}
        \centering
        \includegraphics[width=0.95\linewidth]{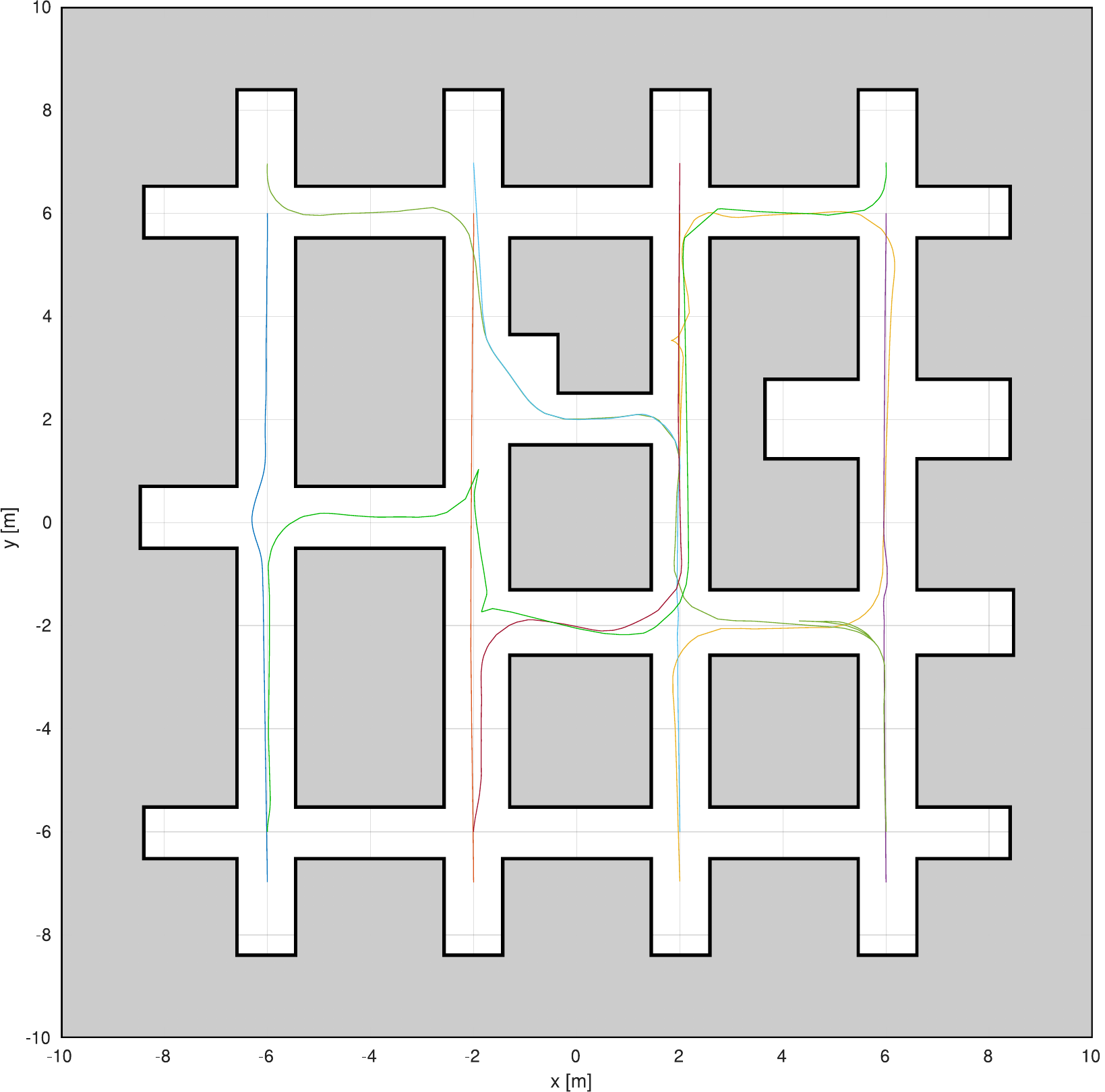}
        \captionsetup{width=.9\linewidth}
        \subcaption{Sample trajectory of eight robots controlled by DTEB Planner in the warehouse map}
        \squeezeup
        \label{fig:warehouse_teb_2}
    \end{subfigure}
    \begin{subfigure}[b]{0.239\textwidth}
        \centering
        \includegraphics[width=0.95\linewidth]{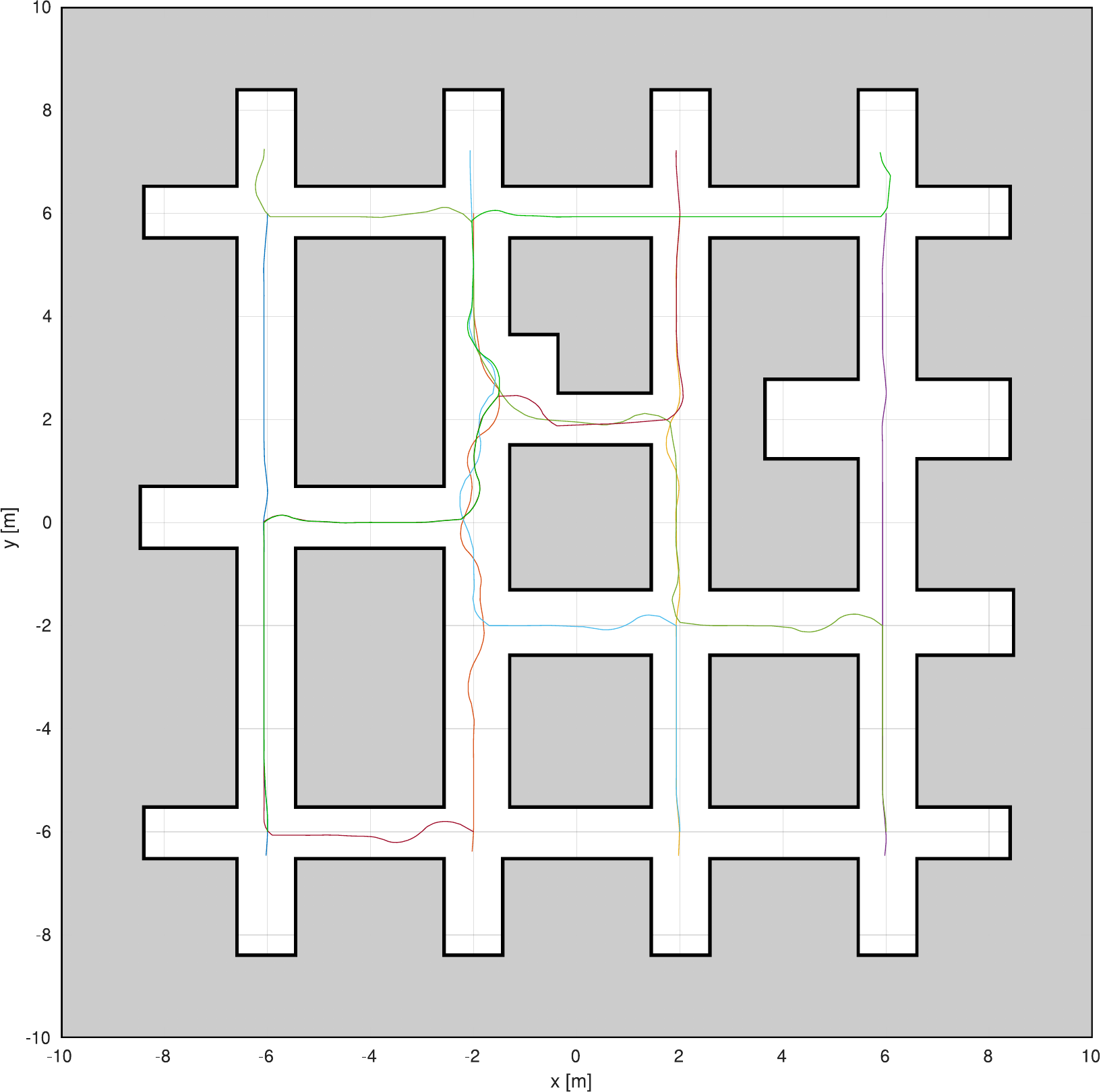}
        \captionsetup{width=.9\linewidth}
        \subcaption{Sample trajectory of eight robots controlled by TUW Route Planner in the warehouse map}
        \squeezeup
        \label{fig:warehouse_tuw_2}
    \end{subfigure}
    \caption{Predictive test using asymmetric goal positions}
\end{figure}
\par

First, we assign the robots' opposite positions as the goals. The proposed planner is able to perform avoidance and wait at the crossroads as shown in \shortautoref{fig:warehouse_teb_1}. 

To add more variety, the test is repeated, but the upper 4 robots use opposite positions as their goals while the lower 4 robots use the diagonal positions as their goals so that the goal positions are asymmetric. Now, there are potential deadlocks involving 3-4 robots in the middle of the map. The proposed planner is still able to handle this scenario without collisions or deadlocks, as shown in \shortautoref{fig:warehouse_teb_2}. 

The TUW Route Planner which also employs Prioritized Planning is able to handle these scenarios. The resultant robot trajectories are shown in \shortautoref{fig:warehouse_tuw_1} and \shortautoref{fig:warehouse_tuw_2}. The comparison between the proposed DTEB Planner and TUW Planner is shown in \shortautoref{tab:comparision_predictive}. 
\begin{table}[!htb]
\caption{Comparison of the results of eight-robot test between TUW route planner and the proposed planner}
\label{tab:comparision_predictive}
\squeezeup
\resizebox{0.48\textwidth}{!}{%
\begin{threeparttable}
\begin{tabular}{l|cc|cc}
\hline
\multicolumn{1}{c|}{}            & \multicolumn{2}{c|}{Opposite Goal Positions} & \multicolumn{2}{c}{Asymmetric Goal Positions} \\\hline
\multicolumn{1}{c|}{} &
  \begin{tabular}[c]{@{}c@{}}Distributed TEB\\ Planner\end{tabular} &
  \begin{tabular}[c]{@{}c@{}}TUW Route\\ Planner\end{tabular} &
  \begin{tabular}[c]{@{}c@{}}Distributed TEB\\ Planner\end{tabular} &
  \begin{tabular}[c]{@{}c@{}}TUW Route\\ Planner\end{tabular} \\
Total travel distance (m) & 110.10 ±3.94             & 117.29 ±0.38            & 152.25 ±7.64         & 143.54 ±0.03        \\
Time to finish (s)      & 22.80 ±2.09             & 43.58 ±5.25       & 44.42 ±4.39        & 47.30 ±0.16         \\
Computation time per robot (ms)      & 20.75 ±9.09             & 4.88 ±0.16*       & 20.13 ±8.95        & 5.04 ±0.26*         \\\hline
\end{tabular}%
        \begin{tablenotes}
          %\footnotesize   %% If you want them smaller like foot notes
          \item[*] Total time to compute the overall offline routing plan instead of an online planning cycle
        \end{tablenotes}
        \squeezeup
\end{threeparttable}
}
\end{table}

For the first task, the \hl{TUW} Route Planner has worse performance, as it takes a detouring path on the left. For the second task, the two planners have similar performance as the task is more complicated. If the task and map become more complicated, TUW Route Planner should outperform the proposed planner, as it has centralized and global information about the planning task. However, since the TUW Route Planner is not an online trajectory/motion planner, it cannot perform reactive avoidance itself.

\begin{figure}[htb]
  \centering
    \includegraphics[width=0.80\linewidth]{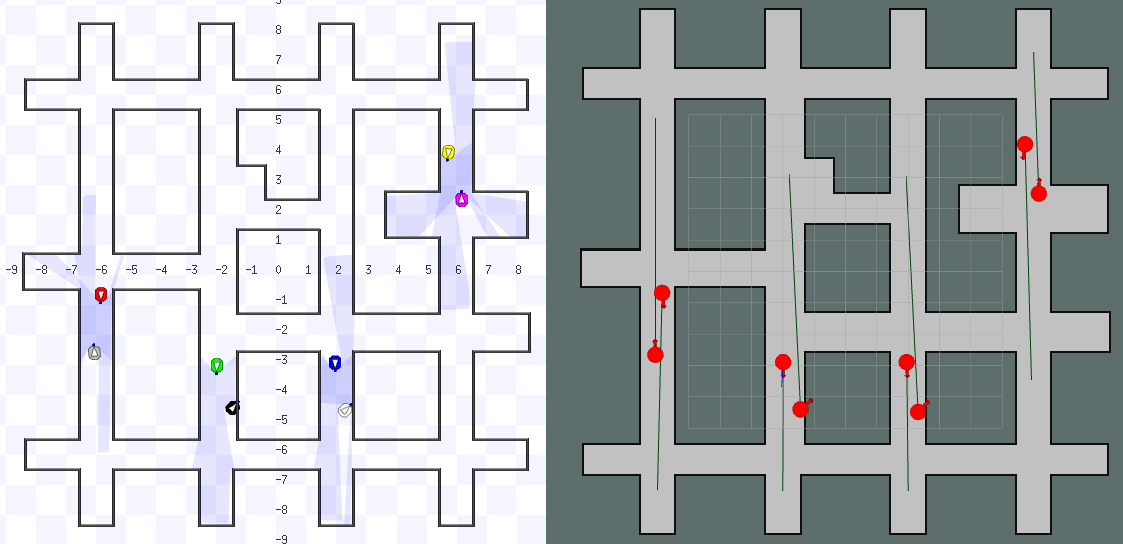}
    \caption{\hl{HRVO Planner is trapped in deadlock in the warehouse map}}
    \label{fig:warehouse_hrvo}
\end{figure}

In comparison to other methods, the HRVO Planner fails to avoid deadlocks in this map. The planner assumes the robots maintain constant velocity for a short period of time and cannot find the solution of avoiding and waiting at a crossroad. The robots will either be stuck in deadlocks or even collide, as shown in \shortautoref{fig:warehouse_hrvo}.

\section{Discussion and Conclusion}

In this paper, we proposed a novel Distributed Timed Elastic Band Planner for multi-robot trajectory planning. The planner combines Prioritized Planning and TEB to allow both reactive and predictive collision avoidance.

Our proposed planner shows better versatility than the previous planners. Reactive planners like HRVO cannot predict long term deadlocks, and route planners like TUW Route Planner cannot perform reactive avoidance. Meanwhile, our proposed planner has a performance similar to or even slightly better than these planners under the discussed environment settings.   
Also, our proposed planner is integrated into ROS navigation stack and has the consideration of robot perception. Beside considering other robots as obstacles using their shared trajectory, the planner can also consider other obstacles perceived by the laser scanner, such as unknown objects on the floor or human operators. The existing pipeline from the single-robot TEB planner is used to track the obstacles.

However, the well-rounded functionality comes at the cost of relatively high computational burden. Although the computational efficiency of the proposed planner is still similar to the original TEB Planner, it is not as efficient as the HRVO or TUW Route Planner, which can control a dozen of robots using a single computer.

Future study can use real world robots to further validate the planner's performance in real-life application, and GPU accelerated optimization can be implemented to further improve the efficiency.

The ROS implementation and test sets are published in GitHub repository \url{https://github.com/chungym/distributed_teb_multi_robot}.

\bibliographystyle{unsrt}
\bibliography{Literature.bib}

\end{document}